\documentclass{article} 
\usepackage[submission]{aaai24}  
\usepackage{times}  
\usepackage{helvet}  
\usepackage{courier}  
\usepackage[hyphens]{url}  
\usepackage{graphicx} 
\usepackage{natbib}  
\usepackage{caption} 
\usepackage{algorithm}
\usepackage{algorithmic}
\usepackage{amssymb}
\usepackage{mathtools}

\usepackage{newfloat}
\usepackage{listings}
\DeclareCaptionStyle{ruled}{labelfont=normalfont,labelsep=colon,strut=off} 
\floatstyle{ruled}
\newfloat{listing}{tb}{lst}{}
\floatname{listing}{Listing}
\title{SMAUG: A Sliding Multidimensional Task Window Based MARL Framework for Adaptive Real-Time Subtask Recognition}
\author{
    Wenjing Zhang\textsuperscript{\rm 1}\equalcontrib, 
    Wei Zhang\equalcontrib
}
\affiliations{
    Harbin Institute of Technology\\
    zhangwenjing763@gmail.com,
    weizhang@hit.edu.cn
}

\usepackage{bibentry}

\begin{document}
\maketitle

\begin{abstract}Instead of making behavioral decisions directly from the exponentially expanding joint observational-action space, subtask-based multi-agent reinforcement learning (MARL) methods enable agents to learn how to tackle different subtasks. Most existing subtask-based MARL methods are based on hierarchical reinforcement learning (HRL). However, these approaches often limit the number of subtasks, perform subtask recognition periodically, and can only identify and execute a specific subtask within the predefined fixed time period, which makes them inflexible and not suitable for diverse and dynamic scenarios with constantly changing subtasks. To break through above restrictions, a \textbf{S}liding \textbf{M}ultidimensional t\textbf{A}sk window based m\textbf{U}ti-agent reinforcement learnin\textbf{G} framework (SMAUG) is proposed for adaptive real-time subtask recognition. It leverages a sliding multidimensional task window to extract essential information of subtasks from trajectory segments concatenated based on observed and predicted trajectories in varying lengths. An inference network is designed to iteratively predict future trajectories with the subtask-oriented policy network. Furthermore, intrinsic motivation rewards are defined to promote subtask exploration and behavior diversity. SMAUG can be integrated with any Q-learning-based approach. Experiments on StarCraft II show that SMAUG not only demonstrates performance superiority in comparison with all baselines but also presents a more prominent and swift rise in rewards during the initial training stage.
\end{abstract}

\section{Introduction}

Cooperative multi-agent reinforcement learning (MARL) has extensive applications, including sensor networks\cite{wu2010multi,shakshuki2009multi,chen2009multi}, robot swarms\cite{huttenrauch2017guided}, urban traffic\cite{cao2012overview,singh2020hierarchical}, and many other fields\cite{mnih2015human,liao2020iteratively,ren2022multi}, and has  significant potential for future development. Compared to single-agent environments, multi-agent systems (MAS) face numerous challenges. Firstly, joint action-value learning requires training a centralized policy reliant on the complete global state. However, as the number of agents increased, the dimensions of the joint state-action space (or the observation-action space) grows exponentially, resulting in ``the curse of dimensionality"\cite {daum2003curse}. Furthermore, due to partial observability and communication constraints, such global information is often difficult to obtain in practical applications. 
To address these issues, independent learning \cite{tan1997multi} is proposed, wherein each agent learns its individual decentralized policy. However, the local observation of each agent is influenced by the behaviors of all other entities in the environment, including ally agents. This dynamic interaction creates a highly unstable environment for MAS. As the number of agents increases, this instability is exacerbated, diminishing the effectiveness of independent learning.

Subsequently, by expanding the paradigm of Centralized Training with Decentralized Execution (CTDE) \cite{foerster2016learning,gupta2017cooperative}, a sequence of value decomposition methods have emerged, such as VDN\cite{sunehag2017value}, QMIX\cite{rashid2020monotonic}, QTRAN\cite{son2019qtran}, and QPLEX\cite{wang2020qplex}, which have attracted considerable attention. However, the majority of value decomposition methods based on the CTDE paradigm just satisfy the sufficiency of the Individual-Global-Max (IGM) principle\cite{rashid2020monotonic} without addressing its necessity, leading to limitations in the function's representation space and approximation capabilities. As a result, these methods converge towards local optima, and their performance remains unsatisfactory\cite{rashid2020weighted}. With the growing number of agents, the solution space of the original problem further expands, exacerbating this issue.
\begin{figure*}[htbp]
    \centering
    \includegraphics[width=0.83\linewidth]{ 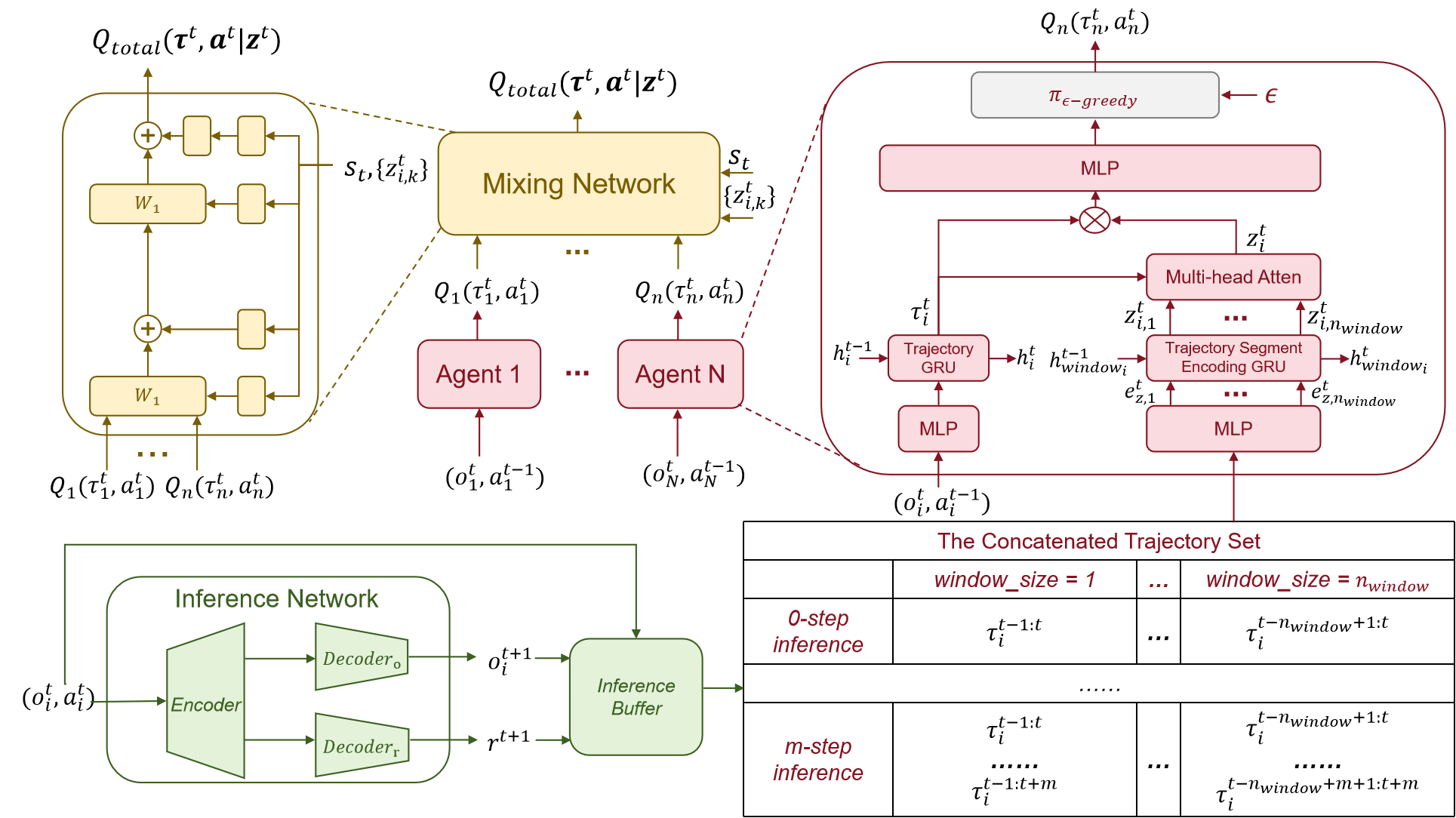}
    \caption{The architecture of SMAUG. The green part denotes the inference network, the red part signifies the subtask-oriented policy network and the yellow part represents the mixing network. The concatenated trajectory set contains the trajectories of different window sizes concatenated with current and predicted trajectory segments. (1) The inference network iteratively infers the observations and rewards of future 1 to \textit{m} time steps based on the subtask-oriented policy network, the current observation, and actions to predict trajectories. (2) The subtask-oriented policy network takes the concatenated trajectory set to produce local action values, $\{Q_i(\tau_i^t,a_i^t)\}$. (3) The mixing network utilizes $\{Q_i(\tau_i^t,a_i^t)\}$ to generate the overall action-value function $Q_{total}(\tau^t,\textbf{a}^t|\textbf{z}^t)$.The mixing network derives its hyperparameters from the current state and subtask set of agents $\{z_{i,k}^{t}\}$}
    \label{fig:1}
\end{figure*}

Inspired by human team cooperation and role-based MARL\cite{wang2020roma,wang2020rode}, subtask-based MARL\cite{yang2022ldsa,yuan2022multi,iqbal2022alma} decomposes a complex task into subtasks. Then, an individual agent can learn to solve distinct subtasks rather than conduct a costly direct exploration in the joint observational-action space. Once tasks are decomposed, the complexity of multi-agent cooperation can be effectively reduced. Task decomposition can effectively divide global tasks into multiple local subtasks, enabling agents to focus and operate more efficiently when solving subtasks.
However, existing subtask-based MARL methods have primarily adopted hierarchical reinforcement learning (HRL) architectures, that each agent can only perform a specific subtask during a fixed period of time\cite{liu2022heterogeneous,yang2019hierarchical}. Besides, some implementations limit the number of subtasks \cite{yang2022ldsa,liu2022heterogeneous,iqbal2022alma,yuan2022multi}. These restrictions may affect the capability of the aforementioned approaches to swiftly respond to abrupt shifts in subtasks or the flexibility of them to define the optimal number of subtasks.

In order to break through the above limitations and facilitate a flexible response to varied and dynamic environments, the subtasks must be dynamically identified and adapted in real-time, and the collaboration of multi-agents should be improved to ensure the successful accomplishment of complex tasks.

To solve the above challenges, we propose a real-time subtask recognition MARL method, called \textbf{S}liding \textbf{M}ultidimensional \textbf{T}ask \textbf{M}ARL \textbf{A}rchitecture (SMAUG). 
SMAUG can dynamically recognize and switch subtasks in real-time, and adapt to diverse and evolving scenarios. 
In SMAUG, the inference network is elaborately designed for subtask recognition, and
the intrinsic motivation rewards are defined to promote subtask exploration and behavior diversity.
At each time step, SMAUG utilizes the set of subtasks for a multi-agent team as input to the mixing network to enhance the rational credit assignments of agents engaged in distinct subtasks. 

Experimental results demonstrate that SMAUG outperforms all baselines on the StarCraft II micromanagement environments\cite{vinyals2017starcraft,samvelyan2019starcraft}.
Moreover, it effectively balances performance and algorithmic stability. 

\section{Preliminaries}
\subsection{Problem Formulation}
Our method considers multi-agent cooperative tasks and utilizes Decentralized Partially Observable Markov Decision Processes (Dec-POMDPs)\cite{oliehoek2016concise} for modeling. 

Dec-POMDPs are represented by a tuple $G=<I, S, A, P, R, \Omega , O, n, \gamma>$, where $I = \{1, 2, ..., n\}$ denotes a finite set of $n$ agents, $S$ is the state space, $A$ is the finite action set and $\gamma$ denotes the discount factor. In a partially observable setting, the observation $o^{t}_{i}\in\Omega $ for agent $i$ is obtained according to the current state  $s^{t}_{i}$  through the observation function $O(s, i)$ at time step $t$. The  history trajectory of agent $i$ is denoted by $\tau_{i}^{t}\in T \equiv (\Omega \times A)^{*}$. At each time step $t$, each agent $i$ selects an action $a^{t}_{i}\in A$ based on its observation $o^{t}_{i}$, forming a joint action $\textbf{a}^{t}\in A^{n}$, leading to the next state $s^{t+1}$ according to the state transition function $P\left(s^{t+1}|s^{t}, \textbf{a}^{t} \right)$, and an  external global  reward $r^{t} = R(s^{t}, \textbf{a}^{t})$ is obtained.

The goal of MARL methods is to learn a joint policy composed of individual policies, i.e. $\pi=(\pi^{1},...,\pi^{n})$ that maximizes the sum of the expectations of the discounted rewards, i.e. $maximize\mathbb{E}[G_{0}]$, where $G_{t}=\sum_{k=0}^{\infty}\gamma ^{k}r^{t+k}$. Thus, the Q-learning methods aim to learn the joint action-value function for the joint policy $\pi$, where $Q^{\pi}(s^{t}, \textbf{a}^{t})=\mathbb{E}_{s^{t+1:\infty}, \textbf{a}^{t+1:\infty}}[\sum_{k=0}^{\infty}\gamma^{k}r^{t+k}|s^{t}, \textbf{a}^{t}]$.
\begin{figure*}[btp]
    \centering
    \includegraphics[width=0.8\linewidth]{ 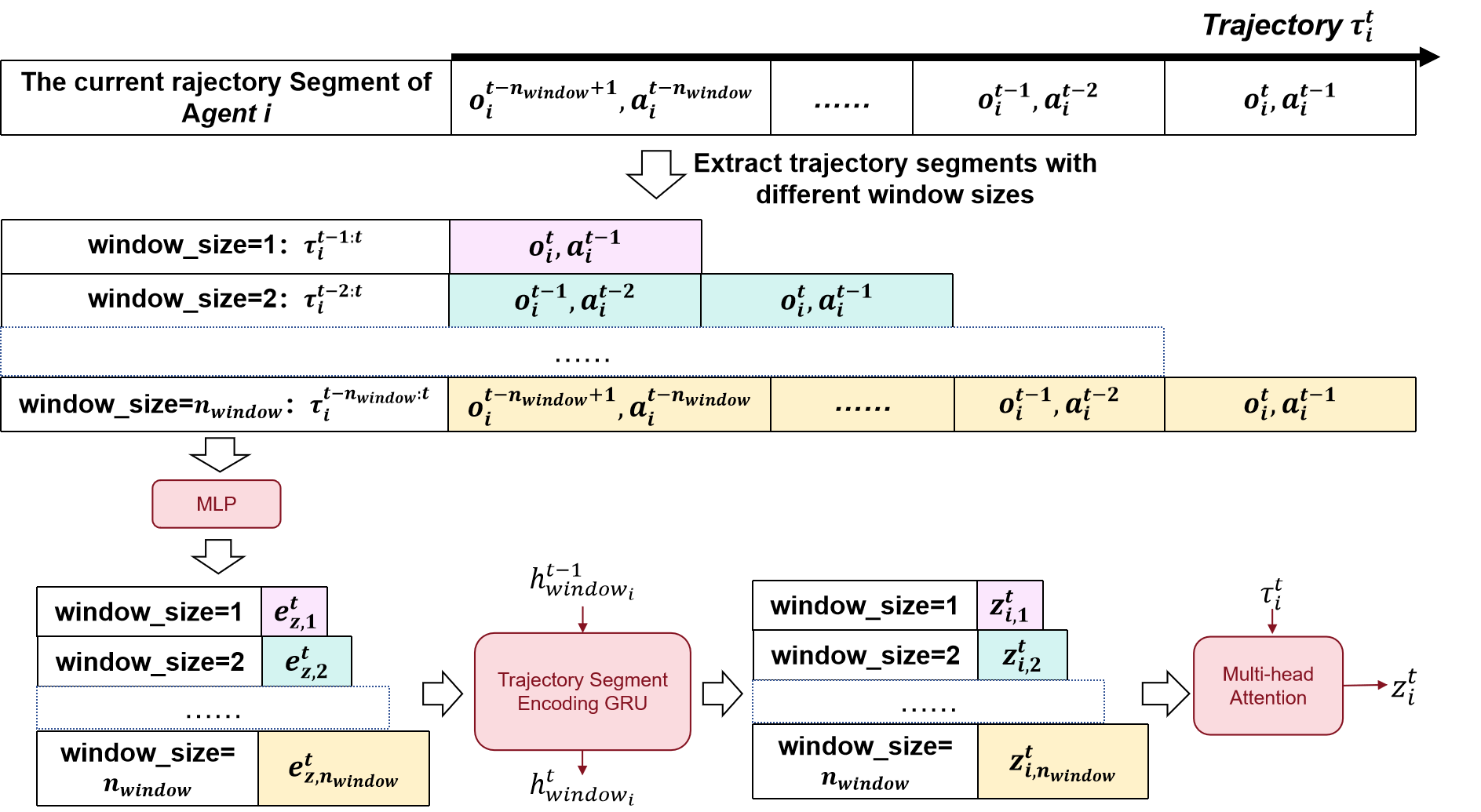}
    \caption{The process of subtask recognition utilizing the sliding multidimensional task window.}
    \label{fig:2}
\end{figure*}
\section{Method}
The architecture of SMAUG is composed of  three components: inference network, subtask-oriented policy network, and mixing network, as Figure \ref{fig:1} presented. The SMAUG framework follows the CTDE paradigm. It learns local action-value functions for agents which are then fed into the mixing network to calculate the global TD loss used for centralized training. The training process of SMAUG includes the below steps : (1) \textit{Subtask Recognition} based on the sliding multidimensional task window, (2) \textit{Subtask Exploration} based on the intrinsic motivation reward function, (3) \textit{Subtask Prediction} based on the inference network and (4) \textit{Subtask-oriented Policy Network Training}.  During execution, the mixing network is removed, and each agent acts based on its locally subtask-oriented policy network with or without the inference network. 

\subsection{Subtask Recognition based on Sliding Multidimensional Task Window}

Intuitively, historical trajectories contain essential information about various subtasks and can be utilized as the data source for subtask recognition. Generally, the time periods of subtasks are various, and can not be accurately predetermined. Thus, SMAUG employs a sliding multidimensional task window to extract and encode subtask information from trajectory segments of varying lengths, utilizes Gate Recurrent Units (GRU\cite{cho2014learning}) and multi-head attention mechanism\cite{wang2020shapley} to recognize current subtasks of agents. Figure \ref{fig:2} illustrates the detailed process of subtask recognition using the sliding multidimensional task window without the inference network.

The sliding windows can effectively capture trajectory information at various levels of granularity. The maximal size of the sliding window $n_{window}$ can be adjusted according to customized requirements. The historical trajectory segment $\tau^{t-k:t}_{i} = (o_i^{t-k},a_i^{t-k-1},\cdot,o_i^t,a_i^{t-1}$) records observations $o_i^t$ and actions $a_i^t$ of agent $i$  from time step $t-k$  to time step $t$. Our objective is to identify current subtasks from the historical trajectory set $\{\tau^{t-k:t}_{i}\}$, where $k=1,\cdot,n_{window}$. A smaller window size focuses on the behavior pattern of short-term  subtasks, while a larger window size captures that of the long-term subtasks. Then, $n_{window}$ different dimensional subtask encodings $\{e_{z,k}^{t}\}$ at the current time step $t$ can be acquired through the sliding window, where $k=1,\cdot,n_{window}$.

For each subtask encoding $e_{z,k}^{t}$, a trajectory segment encoding GRU is employed to capture temporal dependencies and to obtain multidimensional representations of the subtasks $\{z_{i,k}^{t}\}$, where $k=1,\cdot,n_{window}$. In order to generate the trajectory encoding $\tau^{t}_{i}$ at the current time step $t$, a trajectory GRU is utilized. By combining  $\tau^{t}_{i}$ and $\{z_{i,k}^{t}\}$, the multi-head attention module obtains a comprehensive representation of the subtask $z_{i}^{t}$. The multi-head attention mechanism enables effectively recognizing and emphasizing the most informative dimensions of  subtasks. The weighted combination of the subtask representations can be calculated as follows:
\begin{equation}
    z_{i}^{t}=\sum_k{\alpha_{i,k}\cdot{\textbf{v}_{i,k}^{t}}}=\sum_k{\alpha_{i,k}\cdot\textbf{W}_v}\cdot{{z}_{i,k}^{t}}
\end{equation}

The sum of the weights $\sum_k{\alpha_{i,k}}$ is equal to 1, and $\textbf{v}_{i,k}^{t}$ represents the linear transformation result of the subtask representation ${z}_{i,k}^{t}$ through matrix $\textbf{W}_v$. The attention weight $\alpha_{i,k}$ calculates the correlation between the trajectory segment $\tau_{i}^{t}$ and its associated subtask representation $z_{i,k}^{t}$. It can be computed by  the softmax operator:
\begin{equation}
    \alpha _{i,k}=\frac{exp(\lambda \textbf{W}_q\tau_{i}^{t}\textbf{W}_kz_{i,k}^{t})}{\sum exp\lambda \textbf{W}_q\tau_{i}^{t}\textbf{W}_kz_{i,k}^{t}}
\end{equation}
, where $\lambda \in R^+$ is a temperature parameter, set by default to 1; $\textbf{W}_q$ is a query matrix used to transform $\tau_{i}^{t}$ into queries ;and $\textbf{W}_k$ is a shared key matrix used to transform ${z}_{i,k}^{t}$ into keys.

This representation captures the diverse aspects of subtasks, enabling effective recognition of various subtask patterns and variations.

Overall, the sliding multidimensional task window, along with the GRU and multi-head attention mechanism, forms a powerful tool for identifying and encoding subtask information from trajectory segments. It plays a crucial role in SMAUG and contributes to the enhanced performance and adaptability of the MAS. 

\subsection{Subtask Exploration based on Intrinsic Motivation Reward}
In this section, we propose a subtask exploration method based on mutual information and entropy. To enhance the exploration process of subtasks, three main aspects are considered. First, to ensure the sufficient exploration of different subtasks, the diversity of trajectories for different subtasks is maximized. Additionally, to prevent redundancy in subtask concepts, the trajectory information between different subtasks should be as different as possible. Second, we expect the subtask-oriented policy to explore as many environmental states as possible by leveraging previous historical trajectories, leading to diverse observations. Lastly, we aim to maximize the diversity in behaviors among trajectories belonging to different subtasks. Effective subtask exploration requires diverse behaviors to be generated based on current observations associated with specific subtask trajectories.

To address the first objective, we maximize the mutual information $I(\tau;z)$ between trajectories $\tau$ and sub-tasks $z$ to enhance their association. To promote subtask exploration, we aim to maximize the mutual information $I(o;\tau|z)$ conditioned on $z$, to reinforce the correlation between subtask-based trajectories and observations while encouraging observation diversity. Moreover, to increase the diversity of behaviors among trajectories of different subtasks, we maximize the mutual information $I(a;\tau|o)$ conditioned on the current observations $o$. Finally, to encourage diversity in the behaviors of trajectories belonging to different subtasks, we maximize $H(a|o,\tau)$. The maximization objective is shown as Equation \ref{equ:5}. The detailed derivation can be found in the supplementary materials.
\begin{equation}
\begin{split}
r_{MI}= &I(\tau;z)+I(o;\tau|z)+I(a;\tau|o)+H(a|o,\tau)\\
&\geq E_{o,\tau,z}logp(\tau|o,z)-E_{\tau,a}logp(a|o)\\
\end{split}
\label{equ:5}
\end{equation}
We utilize two networks parameterized by $\theta_q$=\{$\theta_\tau$,$\theta_a$\} to approximate probability distributions:
\begin{equation}
    p(\tau|o,z)=Softmax(q_{\theta_{\tau}}(\tau|o,z))
\end{equation}
\begin{equation}
    p(a|o)=Softmax(q_{\theta_{a}}(a|o))
\end{equation}
In summary, the final optimization objective is shown as follows:
\begin{equation}
\begin{split}
    r_{MI}\geq &E_{o,\tau,z}[\beta_{1}logSoftmax(q_{\theta_{\tau}}(\tau|o,z))\\
    &-\beta_{2}logSoftmax(q_{\theta_{a}}(a|o))]
\end{split}
\end{equation}

\subsection{Subtask Prediction based on Inference Network}

For better subtask recognition by the subtask-oriented policy network, we leverage the inference network to predict future observations and rewards, integrating them into the current decision-making process. The process of continuous multi-step prediction can be achieved by an iterative loop between the inference network and the subtask-oriented policy network. During the multi-step prediction process,  the concatenated trajectory set is generated. Consequently, when making decisions at the current time step $t$, the policy network can take into account the concatenated trajectory set, leading to a more comprehensive evaluation of different decision actions. Additionally, we can construct future discounted rewards $r_f =\sum_{m=0}^{n_{f\_step}}\gamma ^{m}r^{t+m}$  using the rewards generated by the inference network, incorporating them as part of the training target for the subtask-oriented policy network, where $n_{f\_step}$ represents the number of steps for inference.
\begin{figure}[!ht]
    \centering
    \includegraphics[width=0.93\linewidth]{ 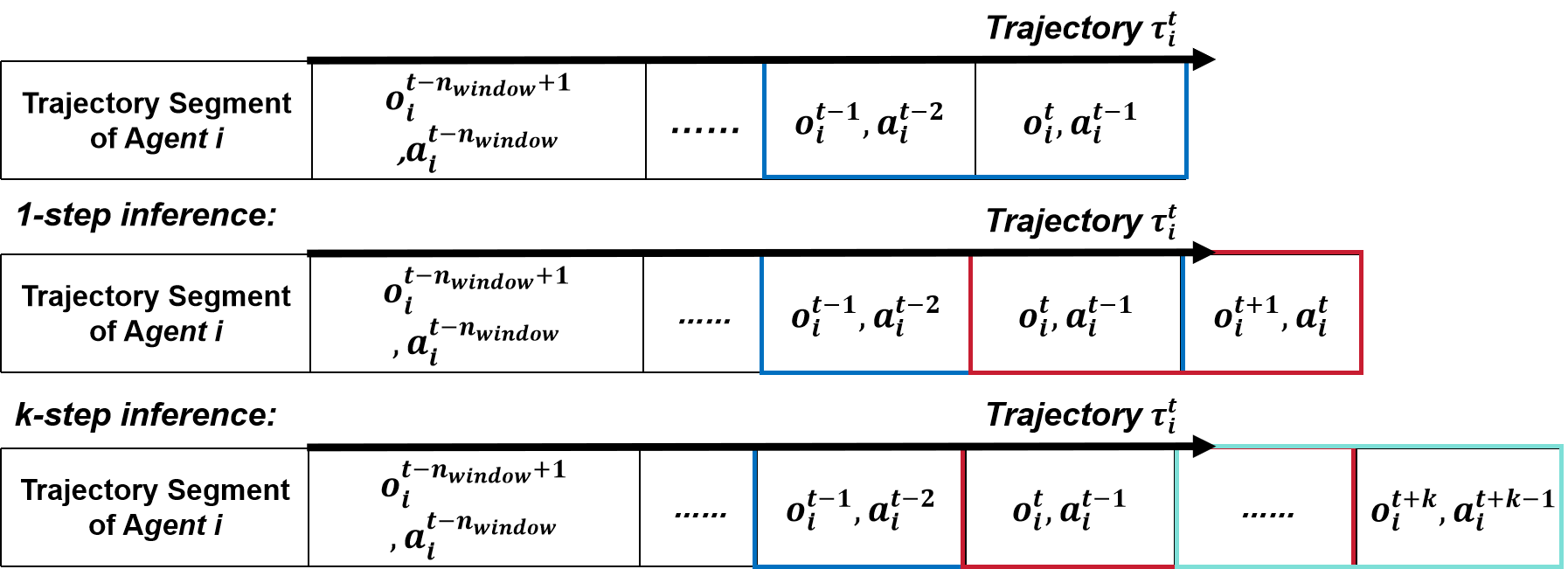}
    \caption{The process of subtask prediction based on the inference network.}
    \label{fig:3}
\end{figure}

The process in Figure \ref{fig:3} illustrates the acquisition of additional subtask trajectory segments through a series of future inferences, utilizing a sliding window with a 2-step size. At time step $t$, the sliding window is positioned at the end of the current trajectory. The first inference involves predicting the observations and rewards for the next step based on the current observation $o_{i}^{t}$ and action $a_{i}^{t}$. The window then slides forward by one time step, and a second inference is conducted to predict the observations and rewards for the following step. This process can be iterated multiple times by an iterative loop between the inference network and the subtask-oriented policy network. By accumulating these inference results of different window sizes, the system obtains the concatenated trajectory set, promoting the subtask-oriented policy network to make more informed actions and better recognize the current subtasks. Such an approach can be particularly useful for tasks that require long-term planning and consideration of future trends.

The inference network consists of a public encoder, a decoder for generating observations, and a decoder for generating rewards. The parameters of the inference network are represented by $\theta_d$. The parameters $\beta_{o}$ and $\beta_{r}$ are used to adjust the weights of the training observation and reward losses, respectively. The training loss function for the inference network is as follows:
\begin{equation}
\begin{split}
    L_{d}=&E[\beta_{o}\sum_{i}\sqrt{(f_{o}(o_{i}^{t},a_{i}^{t})-o_{i}^{t+1})^{2}}+\\&\beta_{r}\sum_{i}\sqrt{(f_{r}(o_{i}^{t},a_{i}^{t})-r^{t+1})^{2}}]
\end{split}
\end{equation}
\subsection{Subtask-oriented Policy Network Training}
\begin{figure}[htbp]
    \centering
    \includegraphics[width=1.0\linewidth]{ 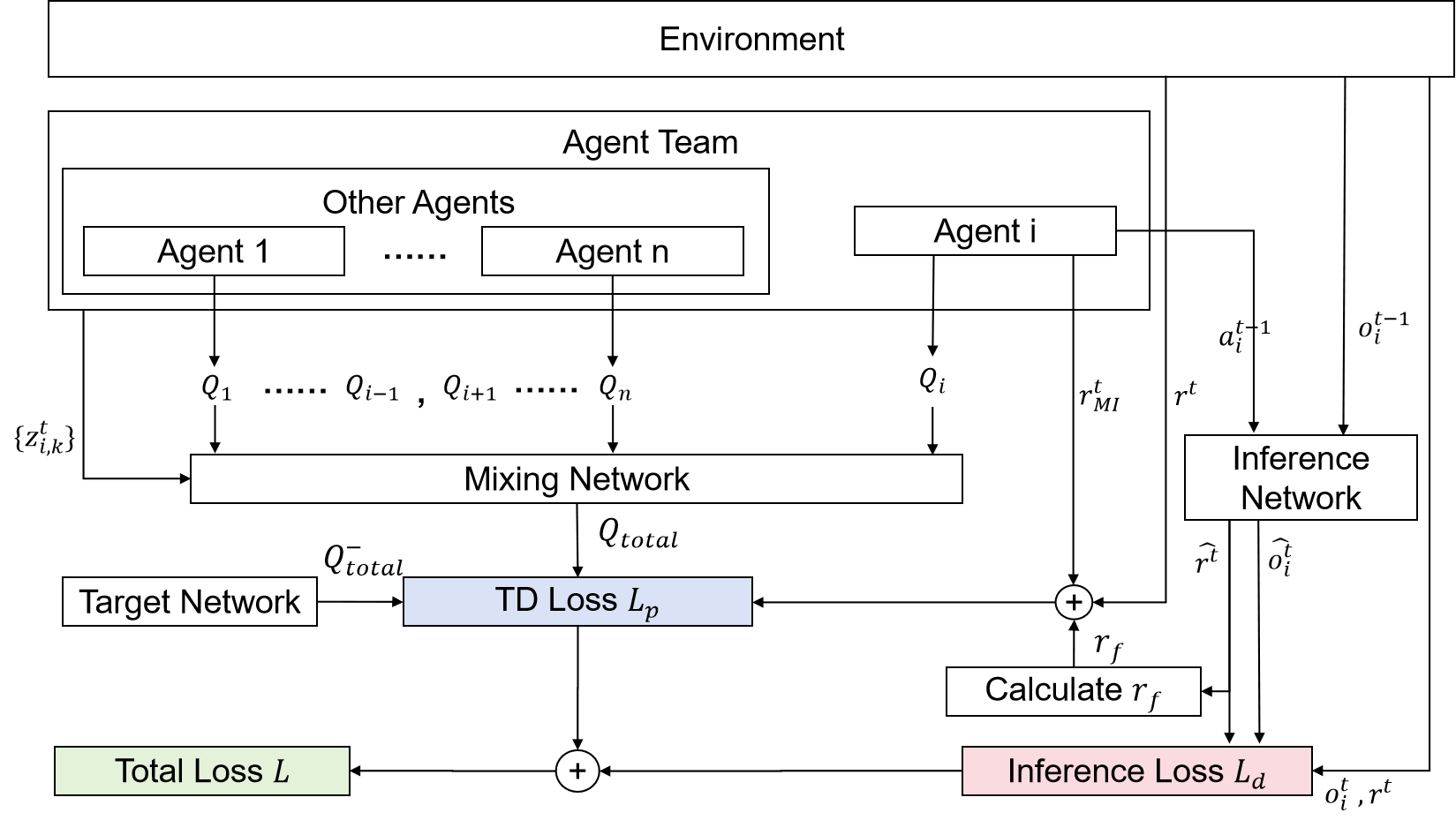}
    \caption{Overall reward design diagram}
    \label{fig:4}
\end{figure}

Finally, we demonstrate how our method steps, including Sub-Task Recognition based on sliding multidimensional task window, Sub-Task Exploration based on Mutual Information, and Sub-Task Prediction based on Inference Network, are integrated into the reward training of SMAUG as shown in Figure \ref{fig:4}.
\begin{figure*}[btp]
    \centering
    \includegraphics[width=0.195\linewidth]{ 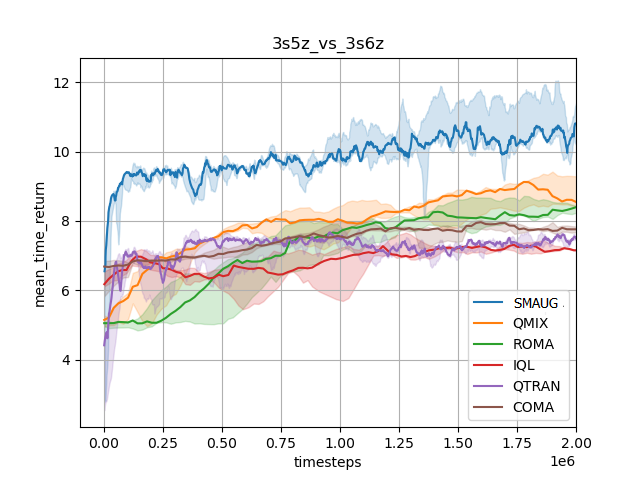}
    \includegraphics[width=0.195\linewidth]{ 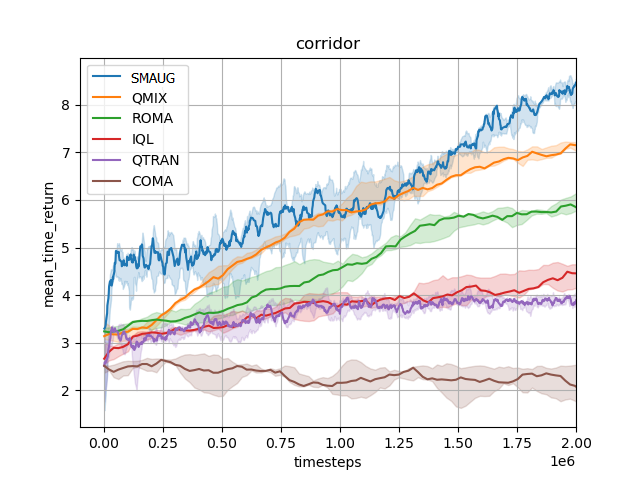}
    \includegraphics[width=0.195\linewidth]{ 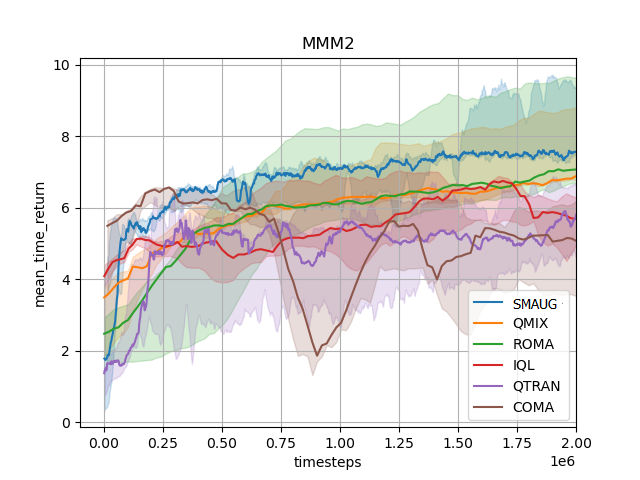}
    \includegraphics[width=0.195\linewidth]{ 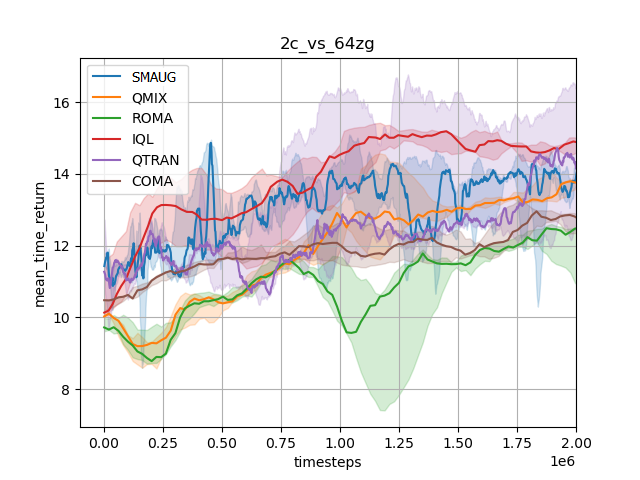}
    \includegraphics[width=0.195\linewidth]{ 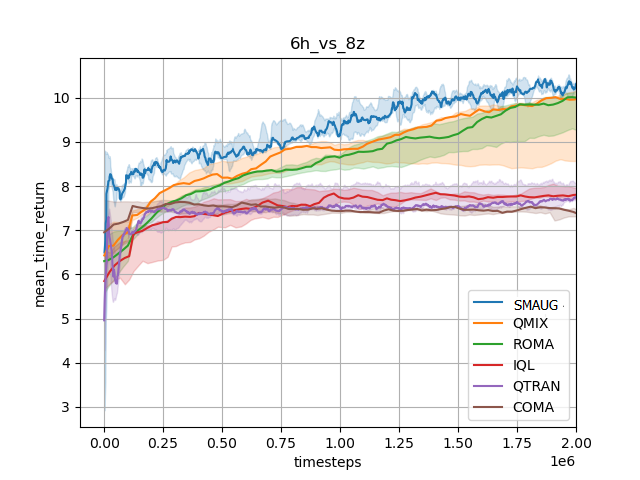}
    \caption{Performance comparison between SMAUG and other baselines in hard maps}
    \label{fig:5}
\end{figure*}

\begin{figure*}[btp]
    \centering
    \includegraphics[width=0.195\linewidth]{ 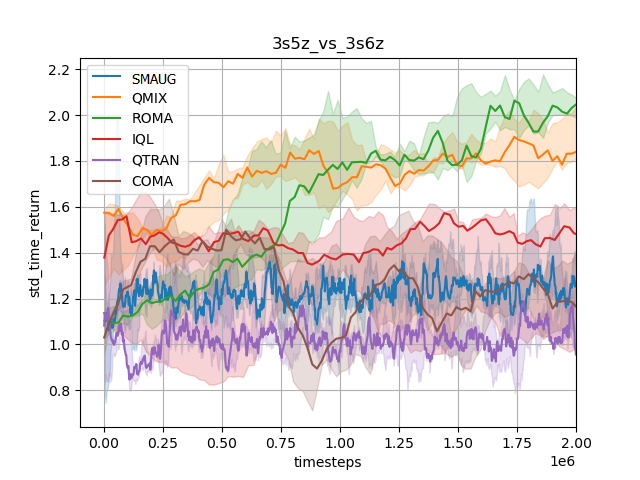}
    \includegraphics[width=0.195\linewidth]{ 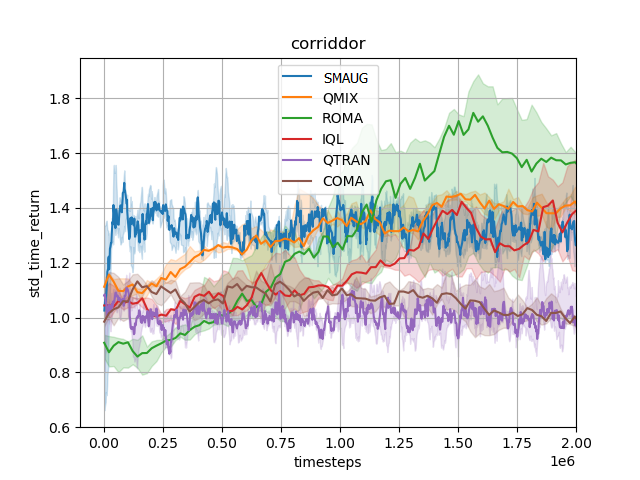}
    \includegraphics[width=0.195\linewidth]{ 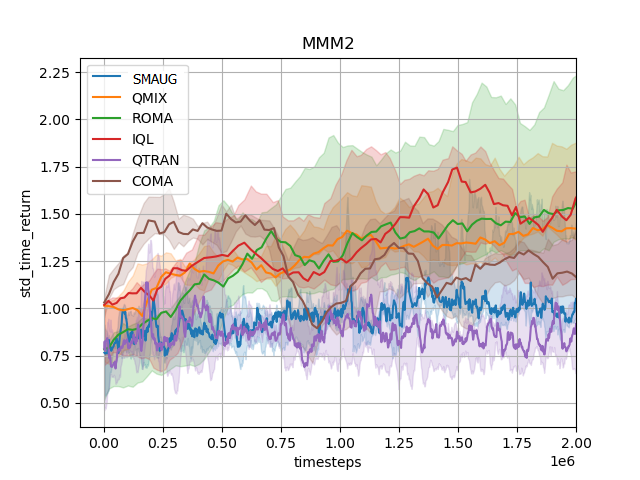}
    \includegraphics[width=0.195\linewidth]{ 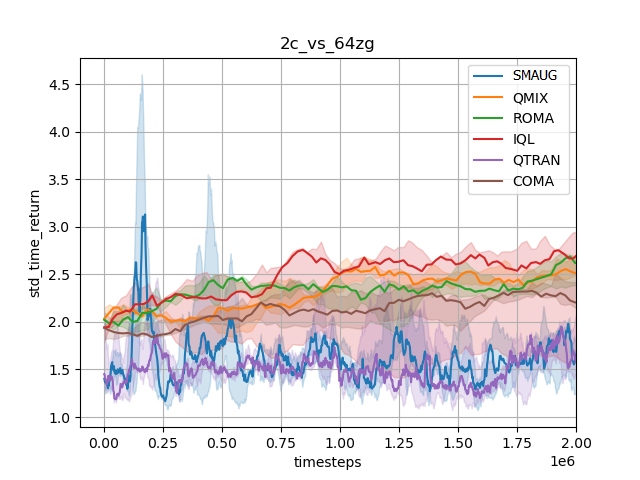}
    \includegraphics[width=0.195\linewidth]{ 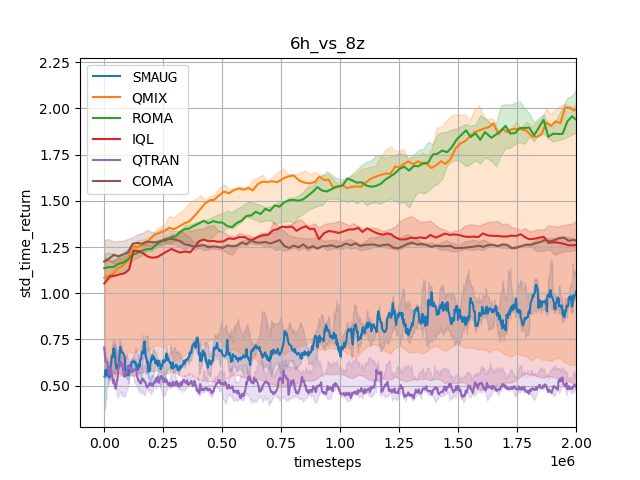}
    \caption{The comparison of standard deviations between SMAUG and other baselines. A smaller standard deviation indicates that the algorithm's performance is more reliable and stable across different situations.}
    \label{fig:6}
\end{figure*}
At each time step $t$, different agents are assigned to execute distinct subtasks. Cooperation or competition among these agents may be necessary to achieve the overall task objective. We combine the representations of different agents' subtasks at the current time $\{z_{i,k}^{t}\}$ to obtain a representation of the overall task at that moment. This representation serves as an abstract depiction of the entire task state. At each time, by utilizing the current state and the subtask set representations $\{z_{i,k}^{t}\}$ as inputs to the QMIX mixing network, we achieve improved weight allocation for agents executing various subtasks. This process guides the QMIX architecture to make trade-offs among different subtasks from a task-level perspective, rather than solely relying on individual agents focusing on local subtask objectives.

The subtask-oriented policy network training uses a QMIX-style mixing network to integrate subtask-oriented action-values  $Q_i(\tau_i^t,a_i^t)$ generated by individual policy networks parameterized by $\theta_p$, resulting in a joint action-value $Q_{total}(\tau^t,\textbf{a}^t|\textbf{z}^t)$ for the entire task. The sub-task-oriented TD loss function is as follows:
\begin{equation}
\begin{split}
&L_{TD}=E[(r^t+\beta_{MI}\cdot r^t_{MI}+\beta_{f}\cdot{r^t_{f}}\\&+\gamma max_{a^{t+1}}Q_{total}^{-}(\tau^{t+1},\textbf{a}^{t+1}|\textbf{z}^{t+1})\\
&-Q_{total}(\tau^t,\textbf{a}^t|\textbf{z}^t))^{2}]
\end{split}
\end{equation}
$r^t$ represents external reward. $r^t_{MI}$ represents mutual information intrinsic reward. $r^t_{f}$ represents future discounted reward at time step $t$.  $Q_{total}^{-}$ represents the total action-value of the target network. $\beta_{MI}$ and  $\beta_{f}$ are used as hyperparameters for the intrinsic reward and future discounted reward, respectively. 

\section{Experiments}
\textbf{Experimental environment }StarCraft Multi-Agent Challenge (SMAC) II\cite{vinyals2017starcraft,samvelyan2019starcraft} is designed to provide a complex and challenging task to test and promote the development of MARL algorithms. In MARL research, the choice of maps can significantly impact the effect among agents, the effectiveness of strategies, and the difficulty of solving tasks. The SMAC II environment offers a set of  StarCraft II maps along with a series of tasks and interfaces suitable for multi-agent systems.

In the environment, each agent is controlled by the policy network and interacts with the environment by taking specific actions. Agents can move in four basic directions, stop, choose an enemy to attack, or do nothing at each time step. Therefore, the action space for each agent consists of  $n_{enemy}$+6 discrete actions, where  $n_{enemy}$ is the number of enemies.

\textbf{Baselines} The SMAC benchmark \cite{samvelyan2019starcraft} comprises 14 maps classified as easy, hard, and super hard. In this study, we  compare with current state-of-the-art value-based MARL algorithms (ROMA\cite{wang2020roma}, QTRAN\cite{son2019qtran}, QMIX\cite{rashid2020monotonic}, COMA \cite{foerster2018counterfactual}, and IQL \cite{tampuu2017multiagent} ) on the super hard maps MMM2, corridor, 3s5z-vs-3s6z, 6h-vs-8z, and the hard map 2c-vs-64zg for comparison. 

\textbf{Hyperparameters} For the purpose of evaluation, all experiments presented in this section are carried out with 5 different random seeds. In the context of all conducted experiments, we set the maximum sliding window size $n_{window}$  to 5, $\beta_{MI}$ to  $5\times10^{-2}$,  $\beta_{f}$ to $10^{-2}$, and the discount factor $\gamma$ to 0.99. The optimization procedure uses RMSprop, employing a learning rate of  $5\times10^{-4}$, $\alpha$ of 0.99, while abstaining from incorporating momentum or weight decay.  In terms of exploration, we have employed an $\epsilon$-greedy strategy, with  $\epsilon$ linearly annealed from 1.0 to 0.05 over $5\times10^4$ time steps. Subsequently, the  $\epsilon$ value remains constant for the subsequent stages of training. Our approach involves operating with 8 parallel environments for the collection of samples. These samples are then organized into batches consisting of 32 episodes, extracted from the replay buffer. These parameter configurations are reminiscent of those employed in QMIX. Furthermore, all experiments are carried out using the computational power of an NVIDIA GTX 1080 Ti GPU.

\subsection{Performance on StarCraft II}
As shown in Figure \ref{fig:5}, in the super hard and hard maps of StarCraft II, the performance of SMAUG is beyond that of other baseline algorithms after $2M$ training steps. Particularly noteworthy is SMAUG's capacity to swiftly and prominently elevate reward values during the initial phases of training, underlining its rapid learning and adaptability. Moreover, SMAUG maintains a commendably low standard deviation across most super hard and hard maps in Figure \ref{fig:6}, indicating that SMAUG exhibits a higher level of reliability and stability which far exceed the baselines.

While ROMA achieves similar performance to the SMAUG algorithm in MMM2 and 6h\_vs\_8z, it stands out with the highest standard deviation across all super hard maps. In contrast, both QTRAN and COMA exhibit lower standard deviations in the super hard and hard maps, yet their overall performance is notably poor.

In conclusion, SMAUG strikes a balance between performance and algorithmic stability. This assertion is evidenced by its superior results in various challenging scenarios and consistently low standard deviations, indicating that SMAUG can facilitate the exploration and resolution of complex tasks, aligning with our expectations for its performance.
\begin{figure}[htbp]
    \centering
    \includegraphics[width=0.48\linewidth]{ 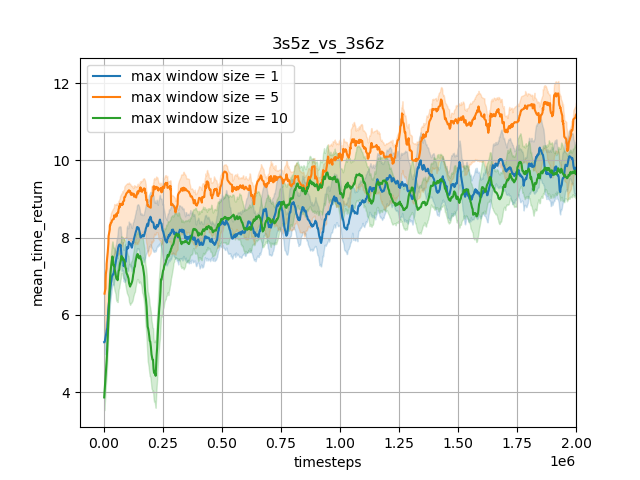}
    \includegraphics[width=0.48\linewidth]{ 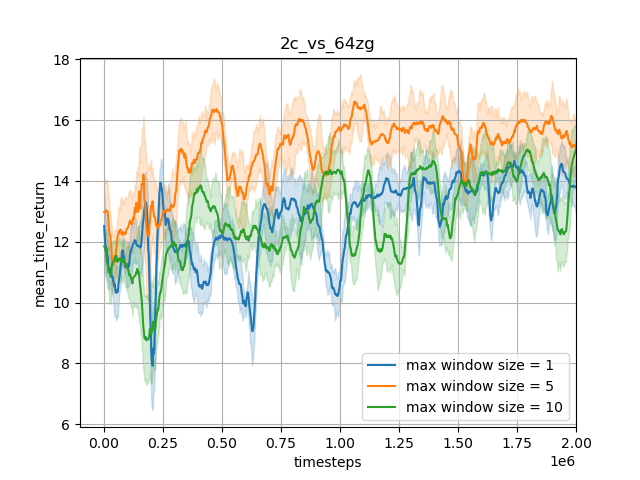}
    \caption{Ablation studies with varying maximum sliding window sizes on 3s5z\_vs\_3s6z (super hard) and 2c\_v\_64zg (hard) maps.}
    \label{fig:7}
\end{figure}
\subsection{Ablation studies}
In this section, we conducted ablation studies across different super hard and hard maps to evaluate the impact of varying maximum sliding window sizes $n_{window}$ on SMAUG. Results are shown in Figure \ref{fig:7}.

We found that in 3s5z\_vs\_3s6z(super hard) and 2c\_vs\_64zg(hard) maps, the experimental results indicate that the best performance is achieved when the maximum window size is set to 5. Interestingly, algorithms with maximum sliding window sizes of 1 and 10 achieve relatively close performance.

This suggests that a balanced window size efficiently captures both short-term and long-term subtask behavior patterns. This balance enhances the decision-making capabilities of the subtask-oriented policy network. A smaller window size (like 1) might mainly focus on immediate behavioral responses, potentially overlooking crucial long-term trends. On the other hand, a larger window size (like 10) may contain more temporal information but could dilute subtask information due to the inclusion of less relevant data points.

These findings emphasize the importance of choosing an appropriate window size to strike a balance between capturing various behavior patterns while maintaining the granularity of subtask information. These insights are crucial for optimizing the performance of the SMAUG algorithm in different scenarios.

\begin{algorithm}[!ht]
\caption{SMAUG}
\label{alg: algorithm}
\textbf{Parameter}: $\theta_p$, $\theta_q$ $\theta_d$
\begin{algorithmic}[1] 
\STATE Initialize parameter vectors  $\theta_p$, $\theta_q$ $\theta_d$\\
\STATE Initialize inference buffer $\hat{D}\gets\{\}$,replay buffer  $D\gets\{\}$, learning rate$\gets\alpha$, window size maximum$\gets n_{window}$, future step$\gets n_{f\_step}$
\FOR{each episode iteration }
\STATE Let $t=0$.
\FOR{each environment step $t$ }
\FOR{$m$ in $n_{f\_step}$ }
    \STATE Obtain action  through the policy network of agent $i$: $a_i^{t+m} \sim \pi_i(a_i^{t+m}|\tau_i^{t+m})$.
    \STATE Obtain observation after one-step inference through the inference network:  $ {{o_i^{t+m+1}} }\sim f_{o_e}(o_i^{t+m+1}|o_i^{t+m},a_i^{t+m})$.
    \STATE Obtain reward after one-step inference through the inference network: $ {r^{t+m}} \sim f_{o_r}(o_i^{t+m+1}|o_i^{t+m},a_i^{t+m})$.
    \IF{m = 1}
    \STATE $\hat{D}\gets \hat{D}\cup\{({o_i^{t+m+1}},a_i^{t+m},{r^{t+m}})\} $.
    \ENDIF
\ENDFOR
 \STATE  Calculate the discounted reward of future steps: $r^t_f = \sum_{j=t}^{t+m}\gamma^{j-t}{r^{j-t}}$.
\STATE Obtain action  through the policy network of agent $i$: $a_i^{t+1} \sim \pi_i(a_i^{t+1}|\tau_i^{t+n_{f\_step}})$.
\STATE Obtain next state : $s^{t+1} \sim P(s^{t+1}|s^{t},\textbf{a}^t)$
\STATE  Obtain next observation: $o_i^{t+1} \sim O(s^{t+1},i)$.
\STATE  Obtain total reward from the environment: $r^t=R(s^{t},\textbf{a}^t)$.
\STATE  Calculate the intrinsic reward: $r^t_{MI}$.
\ENDFOR
\FOR{each gradient step}
    \STATE Update Policy Network and Mixing Network:$\theta_p \gets \theta_p + \alpha\nabla_{\theta_p}L_{TD}$.
    \STATE Update Policy Network by Intrinsic Reward:$\theta_q \gets \theta_q + \alpha\nabla_{\theta_q}L_{TD}$.
    \STATE Update Inference Network:$\theta_d \gets \theta_d + \alpha\nabla_{\theta_d}L_d$.
\ENDFOR
\ENDFOR
\end{algorithmic}
\end{algorithm}

\section{Related Work}
The current landscape of MARL frameworks can be broadly classified into three categories. Firstly, Independent Learning \cite{tan1997multi}, wherein each agent learns decentralized policies. However, this approach often results in instability due to agents treating others as part of the environment. The second category is Joint Action Learning \cite{claus1998dynamics}, which employs centralized policies using complete state information. Yet, partial observability or communication constraints can make global state information unavailable during execution. The third category is the Centralized Training with Decentralized Execution (CTDE) framework \cite{foerster2016learning,gupta2017cooperative}, combining advantages by learning decentralized policies in a centralized manner, improving effectiveness and scalability. Although CTDE algorithms offer solutions for many multi-agent problems, during centralized training, CTDE policies need to search the joint observation-action space, which grows exponentially with the number of agents in MAS. This phenomenon often referred to as the curse of dimensionality \cite{daum2003curse}, leads to challenges in  low sample efficiency, exploration, and computational complexity. Consequently, CTDE algorithms struggle to ensure individual policies converge to global optima.

To address complexity and instability, a decentralized parameter-sharing policy (PDSP) \cite{li2021celebrating} is widely used. It reduces parameters by sharing neural network weights among agents, enhancing learning efficiency. Advanced deep MARL methods use PDSP and CTDE, including value decomposition-based \cite{wang2019learning}, policy gradient-based \cite{lowe2017multi,iqbal2019actor,zhang2021fop}, and other algorithms \cite{foerster2016learning}.

Extending from the PDSP and CTDE, value decomposition methods apply the Individualized Goal Modeling (IGM) principle\cite{rashid2020monotonic} to simplify joint action spaces. However, existing value decomposition-based methods only satisfy the sufficiency of IGM and cannot meet or only partially meet the necessity under certain conditions, limiting the function approximation capacity and resulting in convergence to local optima in most cases. As agent numbers increase, value decomposition becomes inefficient. Recent approaches further decompose and extend the original problem from different perspectives, leading to methods based on roles, skills, and subtasks.

ROMA \cite{wang2020roma} introduces roles to break down the joint action space, allowing agents with similar roles to share experiences and enhance performance. Challenges may arise in distinguishing roles solely from observations during execution. RODE \cite{wang2020rode} decomposes joint action space into role-based local action spaces through action clustering. Challenges may arise due to non-overlapping action sets for roles. HSD\cite{yang2019hierarchical} hierarchically decomposes agent and time dimensions, addressing noisy action-level learning and long-term credit assignment challenges. HSL\cite{liu2022heterogeneous} focuses on distinguishing agents' skills with similar observations, adapting well to scenarios with diverse agent behaviors. However, due to the constraint on the number of subtasks, exhibits a certain lack of flexibility in real-world scenarios. LDSA\cite{yang2022ldsa} improves behavior homogenization issues but is constrained by a fixed number of subtasks, potentially limiting handling dynamic subtask scenarios. MACC\cite{yuan2022multi} introduces task structure decomposability, yet subtask definitions rely on human knowledge and can be overly simplistic. For instance, in  StarCraft II \cite{vinyals2017starcraft,samvelyan2019starcraft}, MACC treats each enemy as a subtask.

\section{Conclusions}
To address the challenges of subtask-based and role-based MARL methods' restrictions, an innovative real-time subtask recognition framework called SMAUG is proposed. The SMAUG framework leverages a sliding multidimensional task window and incorporates a multi-head attention mechanism to construct effective subtask representations. Furthermore, the inference network is designed to assist in subtask recognition, allowing agents to efficiently identify and adapt to varying subtask patterns.
To promote subtask exploration and behavioral diversity in execution, an intrinsic reward function is proposed within the SMAUG framework. 
Experimental evaluations conducted in StarCraft II demonstrate the superiority of SMAUG over value decomposition baselines. Furthermore, SMAUG showcases a higher level of reliability and stability which exceeds all the baselines.

\section{Declaration}
The research idea of this paper was proposed by Wenjing Zhang, who was also responsible for its design, implementation, and experimentation. The writing of the paper was carried out by Wenjing Zhang. Wei Zhang provided the necessary experimental equipment support and contributed to the revision of the paper.

\section{Appendix}

\subsection{Derivation of Intrinsic Reward Functions}
In our approach, the intrinsic motivation reward function based on mutual information is divided into three parts. The first part aims to enhance the diversity of trajectories under different subtasks and to prevent redundancy in subtask concepts. The second part aims to promote diversity in observations under subtask trajectories. The third part aims to encourage diversity in actions under subtask trajectories as well as differences in actions among different trajectories. We revisit our intrinsic motivation reward function and discuss each part separately.

Intrinsic rewards for the diversity of trajectories under different subtasks can be written as:
\begin{equation}
\begin{split}
&I(\tau;z) = E_{\tau,z}log\frac{p(\tau,z)}{p(\tau)\cdot p(z)}
\end{split}
\end{equation}
Intrinsic rewards for the diversity of observations under different subtask trajectories can be written as:
\begin{equation}
\begin{split}
&I(o;\tau|z)=E_{o,\tau,z}log\frac{p(o,\tau|z)}{p(o|z)\cdot p(\tau|z)}
\end{split}
\end{equation}
Intrinsic rewards for the diversity of actions under different subtask trajectories can be written as:
\begin{equation}
\begin{split}
&I(a;\tau|o)+H(a|o,\tau)\\
&=H(a|o)-H(a|o,\tau)+H(a|o,\tau)\\
&=H(a|o)=-E_{\tau,a}logp(a|o)\\
\end{split}
\end{equation}
The overall derivation process is as follows:
\begin{equation}
\begin{split}
&I(\tau;z)+I(o;\tau|z)+I(a;\tau|o)+H(a|o,\tau)\\
&=I(\tau;z)+I(o;\tau|z)+H(a|o)-H(a|o,\tau)+H(a|o,\tau)\\
&=E_{\tau,z}log\frac{p(\tau,z)}{p(\tau)\cdot p(z)}+E_{o,\tau,z}log\frac{p(o,\tau|z)}{p(o|z)\cdot p(\tau|z)}\\
&-E_{\tau,a}logp(a|o)\\
&=E_{\tau,z}log\frac{p(\tau|z)}{p(\tau)}+E_{o,\tau,z}log\frac{p(\tau|o,z)}{p(\tau|z)}-E_{\tau,a}logp(a|o)\\
&=E_{o,\tau,z}logp(\tau|o,z)+H(\tau)-E_{\tau,a}logp(a|\tau)\\
\end{split}
\end{equation}

Because entropy $H(\tau)$ is a positive value, the lower bound of the intrinsic motivation reward function based on mutual information is:

\begin{equation}
\begin{split}
&I(\tau;z)+I(o;\tau|z)+I(a;\tau|o)+H(a|o,\tau)\\
&=E_{o,\tau,z}logp(\tau|o,z)+H(\tau)-E_{\tau,a}logp(a|\tau)\\
&\geq E_{o,\tau,z}logp(\tau|o,z)-E_{\tau,a}logp(a|o)\\
\end{split}
\end{equation}
\subsection{Experimental Details}
\textbf{Baselines}

The baseline methods include value decomposition methods(QMIX\cite{rashid2020monotonic} and Qtran\cite{son2019qtran}), policy gradient-based method (COMA\cite{foerster2018counterfactual}), role-based method (ROMA\cite{wang2020roma}), and independent learning method (IQL\cite{tampuu2017multiagent}). We employ the codes provided by the authors, with their hyper-parameters finely tuned.

\textbf{Architecture}

In this paper, we adopt a QMIX-style mixing network, utilizing the default hyperparameters recommended by the original paper. Notably, we have enhanced the QMIX mixing network by including the current team's subtask set as an additional input component. For individual Q-functions, agents collaborate through a shared trajectory encoding network and a trajectory segment encoding network consisting of two layers: a fully connected layer followed by a GRU layer with a 64-dimensional hidden state. Following these networks, a 16-dimensional multi-head attention module is employed to derive the current subtasks. Furthermore, two separate networks are employed, each incorporating a softmax operator($q_{\theta_{a}}(a|o)$ and $q_{\theta_{\tau}}(\tau|o, a)$), to calculate the lower bound of the intrinsic motivation reward function. While all agents share a one-layer Q network that takes inputs such as the current subtask and trajectory encoding, each agent possesses its independent Q network structured identically to the shared Q network.

For the inference network, we have implemented a shared encoder and two separate decoders to predict the next-step observations and the next-step rewards. The shared encoder is a sequential neural network module composed of linear layers and activation functions. It takes an input and transforms it through hidden layers, applying batch normalization and activation functions to generate an embedding in a lower-dimensional space of size. The decoder consists of a sequential neural network with two linear layers and a ReLU activation function. It takes an input embedding and transforms it through hidden layers to generate outputs. 

\textbf{SMAC Maps}

Next, we will provide an overview of the various maps from the SMAC benchmark on which we conduct experiments.  In the MMM2 map, our team is comprised of 1 Medivac, 2 Marauders, and 7 Marines, while the opposing team is stronger, comprising 1 Medivac, 3 Marauders, and 8 Marines. The corridor map features homogeneous units, with 6 Zealots facing off against 24 enemy Zerglings. Due to the uniformity of enemies, all attack actions result in similar effects. On the map 3s5z\_vs\_3s6z, the player's team includes 3 Stalkers and 5 Zealots, engaging against 3 enemy Stalkers and 6 enemy Zealots. This scenario includes heterogeneous enemy units, leading to distinct outcomes when attacking Stalkers and Zealots. In the scenario 6h\_vs\_8z, 6 Hydralisks confront 8 Zealots. Lastly, the 2c\_vs\_64zg scenario involves only 2 Colossi as allied agents, facing a staggering 64 Zergling enemy units, which is the largest number in the SMAC benchmark. This setting presents a significantly expanded action space for the agents compared to other scenarios.

\end{document}